\documentclass[a4paper,11pt]{article}
\usepackage[a4paper,margin=1in]{geometry}
\usepackage[T1]{fontenc}
\usepackage{graphicx}
\usepackage{tabularx}
\usepackage{linguex}
\usepackage{mathpazo}
\usepackage{inconsolata}
\usepackage{booktabs}
\usepackage{amsmath}
\usepackage[numbers,sort&compress]{natbib}
\usepackage{url}
\usepackage[table]{xcolor}
\usepackage{multirow}
\usepackage{makecell}
\usepackage{titlesec}
\usepackage{lipsum}
\usepackage{etoolbox}
\usepackage[hidelinks]{hyperref}
\usepackage{setspace}
\usepackage{lineno}
\usepackage{fancyhdr}
\usepackage{moreverb}
\usepackage{xspace}
\usepackage{tcolorbox}
\usepackage{enumitem}

\makeatletter
\patchcmd{\@maketitle}{\LARGE}{\Large}{}{}
\makeatother

\titleformat{\section}
  {\normalfont\large\bfseries}{\thesection}{1em}{}

\titleformat{\subsection}
  {\normalfont\normalsize\bfseries}{\thesubsection}{1em}{}

\titleformat{\subsubsection}
  {\normalfont\normalsize\bfseries}{\thesubsubsection}{1em}{}

\title{To model human linguistic prediction, make LLMs less superhuman}

\author{Byung-Doh Oh$^{1,}$\footnotemark \enspace and Tal Linzen$^{2,3}$}

\date{}

\begin{document}
\pagestyle{fancy}
\fancyhead{} % clear all header fields

\setlength{\footskip}{24pt}

\pagenumbering{arabic}

\maketitle

\footnotetext[1]{Division of Linguistics and Multilingual Studies, Nanyang Technological University, Singapore}
\footnotetext[2]{Department of Linguistics, New York University, New York, USA}
\footnotetext[3]{Center for Data Science, New York University, New York, USA}
\renewcommand*{\thefootnote}{\fnsymbol{footnote}}
\footnotetext[1]{Correspondence: \url{byungdoh.oh@ntu.edu.sg} (B.-D. Oh)}

\section*{Abstract}
When we read, we make predictions about upcoming words; these predictions influence our reading behavior.
The success of large language models (LLMs), which, like humans, make predictions about upcoming words, has motivated their use as models of human linguistic prediction.
Surprisingly, in the last few years, as LLMs' ability to predict the next word has improved, their ability to explain reading behavior has declined.
We argue this is because current LLMs can predict upcoming words much better than human readers can. This `superhumanness' is driven by LLMs' extensive training data, stronger long-term memory of training examples, and stronger short-term memory.
We advocate for LLMs with human-like memory and for new experiments to measure the alignment between humans and LLMs, and outline directions towards achieving these goals.

\noindent \textbf{Keywords}: language $|$ prediction $|$ reading $|$ language comprehension $|$ large language models

\section*{LLMs as potential models of prediction in human language comprehension}
Prediction is a core principle of human cognition, where it plays a central role in domains ranging from visual object recognition to action planning \citep{bar_2009_proactive, clark_2013_whatever}. Language comprehension, in particular, is arguably as rapid and efficient as it is due to its predictive nature: When we encounter predictable words, we are able to read them faster, and our brains display reduced neural activity in response to those words \citep{ehrlich_1981_contextual, kutas_1984_brain, smith_2013_effect}.

Studying how linguistic predictions are generated and how they shape language comprehension can help shed light on human cognitive processes in language and beyond. Word predictability estimates from large language models \citep[LLMs;][]{jozefowicz_2016_exploring, gulordava_2018_colorless, radford_2019_language}---neural networks that, like humans, make predictions about upcoming words---could be immensely useful for this project \citep{shain_2024_large, huang_2024_large}.
But for LLMs to serve this purpose,  their word predictions need to align with those made by humans. In the last few years, it has become clear that the predictions made by mainstream LLMs are in fact increasingly diverging from those of humans: LLMs are much better able to predict the next word than humans are; in other words, their predictions are superhuman \citep{oh_2022_comparison, oh_2023_surprisal, shain_2024_large}. In many artificial intelligence applications, superhumanness is benign and even desirable; that is not, of course, the case for cognitive modeling.

This paper highlights two sets of issues that contribute to the growing discrepancy between humans and LLMs.
The first is LLMs' long-term memory: They can consume and retain massive amounts of data, and consequently they know dramatically more than human readers.
The second is their near-perfect short-term memory of previous words in the text: Unlike humans, LLMs can faithfully recall specific words that occurred many pages before the current word.
After discussing these issues, we outline modeling efforts toward developing more human-like LLMs, and argue that, because the empirical data that measures this discrepancy is currently very partial, new human experiments will be necessary to benchmark research progress on this front.

\section*{How LLMs are used to study human language comprehension}

While it is fairly clear that prediction plays a role in language comprehension \citep{kuperberg_2016_prediction, staub_2025_predictability}, there are many open questions about this process.
What kinds of information about the first few words of a sentence do we draw on to predict upcoming words, and what partial representations of the sentence do we construct?
What aspects of the behavioral processes involved in language comprehension, such as eye movements during reading, can be explained by prediction, and what aspects need to be explained by other principles \citep{vanschijndel_2021_single}?

Any quantitative study of this process relies on estimates of how predictable a word in a particular context is to readers (Table \ref{tab:predictability}):
For example, given the partial sentence \textit{I purchased a} \underline{\hspace{1cm}}~, how much more predictable is \textit{banana} compared to \textit{cassowary}?
Traditionally, predictability was estimated using responses to the cloze task \citep{taylor_1953_cloze}, where subjects are asked to provide a completion (usually the next word) given an incomplete sentence; words that are more frequently produced by subjects in a particular context are deemed to be more predictable in that context.
But this method is not very practical: To reliably estimate a difference in predictability between two low-predictability words, which are likely to be produced only rarely in a cloze experiment, enormous samples from many millions of participants are required.

\newcolumntype{L}{>{\raggedright\arraybackslash}X}
{
\renewcommand{\arraystretch}{2}
\begin{table}[t!]
    \vspace{\baselineskip}
    \centering
    \begin{tabularx}{\textwidth}{|>{\hsize=.2\hsize}L|>{\hsize=.4\hsize}L|>{\hsize=.4\hsize}L|} \hline
    \cellcolor{lightgray}Method & \cellcolor{lightgray}Definition & \cellcolor{lightgray}Limitations \\ \hline
    Cloze task \citep{taylor_1953_cloze} & Human subjects are asked to provide a completion to an incomplete sentence such as \textit{I purchased a} \underline{\hspace{1cm}} & Many responses are required to differentiate between low-predictability words, raising concerns about feasibility \\ \hline
    $n$-gram language models \citep{jelinek_1975_design, baker_1975_dragon} & Conditional probabilities are calculated based on sequence counts from a corpus, e.g.~$\displaystyle \frac{c(\textit{I purchased a banana})}{c(\textit{I purchased a})}$ & Some sequences may not be attested in the corpus, and the context window is too short to model human linguistic prediction \\ \hline
    Large language models \citep{jozefowicz_2016_exploring, gulordava_2018_colorless, radford_2019_language} & Conditional probabilities are calculated from a large neural network that is typically trained to predict the next word & The probabilities may be based on long-term and short-term memory capacities that far exceed those of humans \\ \hline
    \end{tabularx}
    \caption{Methods for estimating word predictability and their limitations.}
    \label{tab:predictability}
\end{table}
}

An alternative to the cloze task relies on conditional probabilities derived from $n$-gram language models \citep{jelinek_1975_design, baker_1975_dragon}, which estimate the probability of a word in a context of a handful of preceding words (typically one to four), based on the number of times it occurred in this context in a training corpus.
While predictability estimates derived in this way have substantial predictive value for human word-by-word reading \citep{mcdonald_2003_low, boston_2008_parsing, fossum_2012_sequential, smith_2013_effect, shain_2019_large},
$n$-gram language models are too impoverished to model human linguistic prediction: The context window they consider is far shorter than the context length to which humans are sensitive when making predictions \citep{fitzsimmons_2013_fast, brothers_2020_going}, and the ability of these models to generalize to new contexts that were not seen in the training corpus is very limited.

\begin{figure}[t!]
    \centering
    \includegraphics[width=\linewidth]{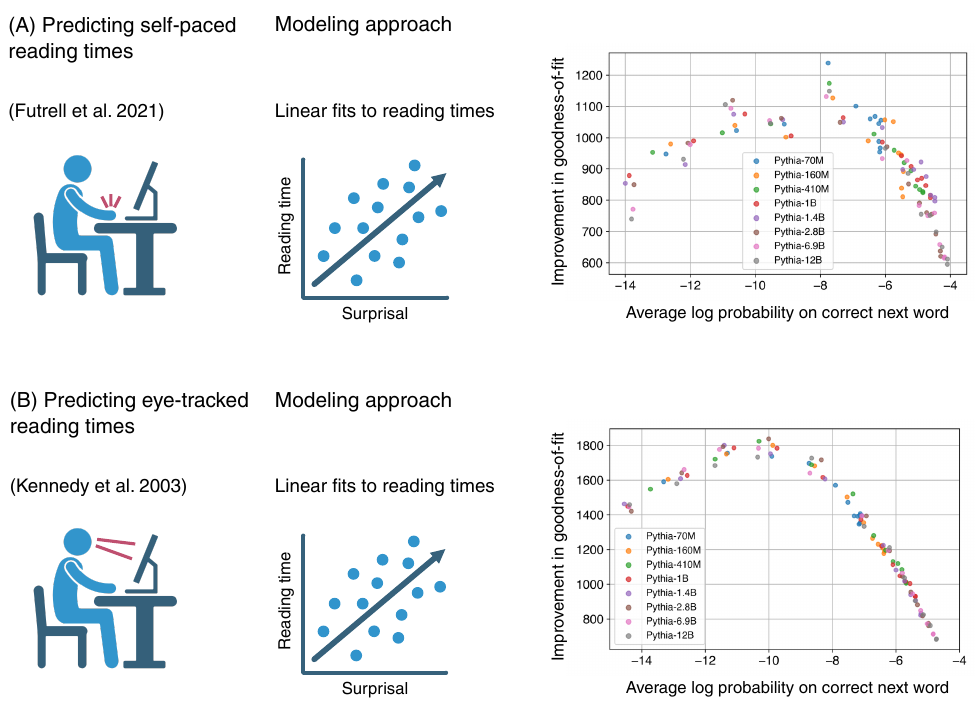}
    \caption{\textbf{Relationship between an LLM's word prediction accuracy and fit to human data.} (\textsf{A}) Using the Natural Stories Corpus \citep{futrell_2021_natural}, which contains self-paced reading times of expository articles and short stories, we reproduce the analysis of \citep{oh_2023_transformer} by calculating word-by-word predictability estimates from a wide range of LLMs \citep{biderman_2023_pythia} and evaluating their linear fit to reading times as the improvement in regression model log-likelihood over a common set of baseline predictors. Each dot represents one LLM; dots with the same color are LLMs with the same number of parameters, but different amounts of training data. The improvement in word prediction accuracy initially improves an LLM's alignment with human data, which is consistent with earlier studies \citep{goodkind_2018_predictive, wilcox_2020_predictive, merkx_2021_human}. In contrast, after a certain point, further improvement in word prediction accuracy drives a misalignment with human data, which consistent with later studies that used the same dataset \citep{oh_2022_comparison, oh_2023_surprisal, shain_2024_large}. (\textsf{B}) The same pattern of results is observed with data from the Dundee Corpus \citep{kennedy_2003_dundee}, which contains eye-tracked reading times of newspaper editorials.}
    \label{fig:trend}
\end{figure}

LLMs \citep{jozefowicz_2016_exploring, gulordava_2018_colorless, radford_2019_language}, which predict upcoming words using neural networks, address both of these issues to a large extent, as the probabilities they produce are determined by longer sequences of words, and their representations can generalize better to new contexts.
These models were quickly adopted by cognitive scientists, who showed that predictability estimates derived from these models are superior to those from $n$-gram models \citep{goodkind_2018_predictive, wilcox_2020_predictive, merkx_2021_human}: They explained more variance in word-by-word reading times, measured as the time taken between keystrokes in an experiment where each keystroke reveals the next word \citep[self-paced reading;][]{just_1982_paradigms}, or as the duration of eye fixations on each word collected through eye tracking \citep{rayner_1998_eye}.
A consistent finding from these earlier studies was that language models that predict the next word more accurately---i.e.~place higher probability on the word that in fact occurred next in the sentence---also yielded conditional probabilities that align more closely with human reading times.

Intriguingly, this relationship completely reversed in studies published more recently (from around 2022 on).
These studies---which used more powerful language models, typically based on the Transformer neural network architecture \citep{vaswani_2017_attention} and trained on a much larger amount of data---showed that these models find the words they encounter to be more predictable than humans do, resulting in a growing divergence between the models' predictions and human reading times \citep[Figure \ref{fig:trend};][]{oh_2022_comparison, oh_2023_surprisal, shain_2024_large}.
This suggests that these newer models---which we will call ``mainstream LLMs'' throughout this paper---make predictions that average human readers readily cannot, making them superhuman as models of linguistic prediction.
We argue that this superhumanness is best characterized by two underlying issues: LLMs' long-term memory of massive amounts of training data and their near-perfect short-term memory of previous words in the text.

\section*{Superhuman long-term memory}
\begin{figure}[t!]
    \centering
    \includegraphics[width=\linewidth]{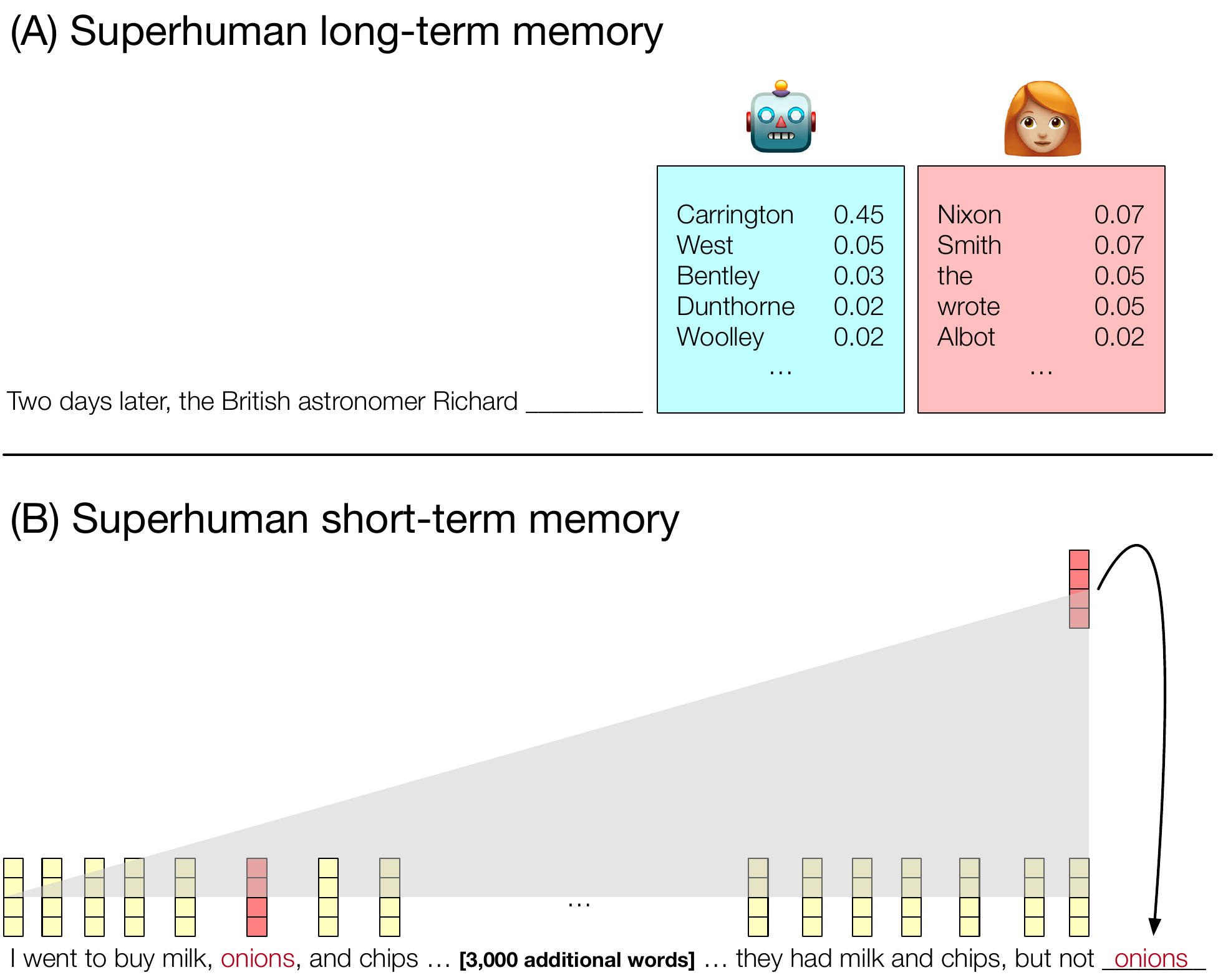}
    \caption{\textbf{The memory capacity of LLMs far exceed those of humans.} (\textsf{A}) Mainstream LLMs have much more factual knowledge than average humans due to their superior long-term memory. Top five words with highest probabilities from the Llama 3 LLM \citep[blue;][]{grattafiori_2024_llama} and proportions of top five frequent completions collected from 41 human subjects \citep[red;][]{luke_2018_provo}. The LLM places high probabilities on names of astronomers, including the correct continuation \textit{Carrington}. (\textsf{B}) Transformers have perfect store of the input sequence. Even with thousands of words in between, Transformers can retrieve the earlier occurrence of the word \textit{onions} to make the correct prediction of the second \textit{onions}.}
    \label{fig:issues}
\end{figure}

A core reason that LLMs exceed the ability of human readers to predict the next word is their ability to remember examples from the training data much more faithfully than humans can. This is particularly clear in cases where successful prediction relies on factual knowledge: The fact that readers bring their background knowledge to bear on the comprehension process has been emphasized in theories of reading comprehension \citep{kintsch_1998_comprehension, dijk_1983_strategies} and studied as a source of individual difference in language comprehension \citep{smith_2021_role}.
Consider the partial sentence \textit{Elvis Presley was born in the city of} \underline{\hspace{1cm}}~, for example.
How predictable the upcoming word is for the reader will crucially depend on whether they already know this fact about the birthplace of Elvis. If they do, they will be more likely to correctly predict \textit{Tupelo} and experience less difficulty upon encountering this word.
But many readers quite plausibly will have never been exposed to information about the birthplace of Elvis Presley, or, if they have been exposed to it at some point, may have completely forgotten it (see Figure~\ref{fig:issues} for a related example).

Given the central role of background knowledge in next-word prediction, an accurate cognitive model of linguistic prediction would need to make predictions that are consistent with the reader's knowledge.
But mainstream LLMs acquire far more knowledge than humans could ever acquire.
This discrepancy can be attributed to three reasons.
First, the data typically used to train mainstream LLMs contains a substantial amount of material that conveys factual information, such as Wikipedia articles.
Second, the amount of data used to train mainstream LLMs is often orders of magnitudes larger than what humans are exposed to \citep{hart_1995_meaningful, wilcox_2025_bigger}: A typical English-speaking child will have experienced at most 100 million words by the age of 12, while mainstream LLMs like \mbox{Llama 3} \citep{grattafiori_2024_llama} are trained on as many as 15 \textit{trillion} words.
Finally, and perhaps most fundamentally, mainstream LLMs do not easily forget the text that they encounter, unlike humans who naturally do so over time.
Mainstream LLMs have been shown to store long sequences of words they have encountered such that it is possible to extract them verbatim by prompting the model \citep{mccoy_2023_much, carlini_2023_quantifying, merrill_2024_evaluating}.
To return to our running example, the properties of the training data make it more likely for mainstream LLMs to be exposed to the birthplace of Elvis Presley, and the learning behavior of these models makes it likely that this piece of knowledge will be retained.
As a consequence, mainstream LLMs become much more likely than human readers to correctly predict \textit{Tupelo} given the example above.

Supporting this conjecture, recent studies provide preliminary correlational evidence that the mismatch between LLM and human linguistic prediction is in part due to the LLMs' superior long-term memory.
A regression analysis of reading times showed that LLMs were much less surprised than human readers by proper names like \textit{Tupelo}, which can only be predicted with accurate factual knowledge \citep{oh_2023_surprisal}.
The discrepancy with humans was more severe for larger, more powerful models, suggesting that improving the next-word prediction accuracy of language models will not make them more human-like---if anything, the opposite is the case.

Another, more indirect source of evidence for this hypothesis comes from the finding that language models trained on smaller amounts of data than typical mainstream LLMs yield probabilities that are better aligned to human reading times \citep{oh_2023_transformer}.
Although limiting the amount of training data could affect language model probabilities in many different ways \citep{kaplan_2020_scaling, chang_2024_characterizing}, one of them is by preventing the exposure of factual information like the birthplace of Elvis that could eventually be memorized.
Further support for this hypothesis comes from the finding that while training language models on some amount of text from specialized domains such as biology and physics textbooks improves their ability to model the reading behavior of subjects with the relevant domain expertise, training on too much data from those domains has a deleterious effect, suggesting again that the models acquire superhuman factual expertise \citep{skrjanec_2026_language}.

We hypothesize that the mismatch in the familiarity with the text between humans and LLMs goes beyond factual, knowledge-based prediction. Consider, for example, multiword expressions like the idiom ``spill the beans,'' where the later words are easy to predict given the first ones \citep{rambelli_2023_frequent}. Mainstream LLMs are likely able to store a much larger range of such expressions from different dialects and registers of the language than humans can, thanks to their vast training data and strong memorization capabilities. Consequently, they are likely to predict upcoming words in these expressions with greater accuracy than humans.
An extreme case of this issue could be illustrated by famous documents like the Declaration of Independence, which are likely to be memorized and predicted accurately by mainstream LLMs \citep{mccoy_2023_much, merrill_2024_evaluating}, but are not likely to be memorized by most humans.
Crucially, the amount of exposure required for mainstream LLMs to become familiar with these examples is much lower than that required for humans: In some cases, LLMs can memorize a training example after only a handful of exposures \citep{tirumala_2022_memorization, lesci_2024_causal}.

In summary, a crucial factor that limits the ability of LLMs to serve as models of linguistic prediction is their greater familiarity with facts and turns of phrase compared to the average human reader.
This is due to the difference in the quantity and quality of language exposure and the ability to remember what was seen.
As a consequence, mainstream LLMs embody massive amounts of information that enable them to predict upcoming words that are unpredictable to humans with limited knowledge.

\section*{Superhuman short-term memory}
Whether it is spoken or written, linguistic input is transient during language processing.
From this impermanent input, the goal of the human comprehender is to build a mental representation of its meaning \citep{bransford_1971_abstraction, jarvella_1971_syntactic}.
During this process, humans are subject to limitations in working memory and naturally forget parts of the input over time \citep{baddeley_1974_working, baddeley_2003_working, lewis_2005_activation}.
The decay in the availability of earlier linguistic input influences readers' expectations about upcoming words \citep{futrell_2020_lossy, hahn_2022_resource}.

In contrast to humans, mainstream LLMs are able to build and maintain robust representations of a much larger number of previous words and use these representations to make next-word predictions.
Part of this is attributable to the Transformer neural network architecture \citep{vaswani_2017_attention} that underlies most mainstream LLMs at the time of writing.
These networks build vector representations of each word by taking a weighted average of the representations of each previous word in the sequence (this is referred to as ``attention'').
During this process, the model has access to its representations of the words in a very large context window which can span tens of thousands, or even millions of words---typically enough to fit the entire length of the text being processed. The resulting representation of the sequence is not subject to any form of temporal decay, unlike human memory.

Mainstream LLMs' substantially stronger short-term memory could allow them to make more accurate predictions. Names and concepts are much more likely to be mentioned again once they have already been mentioned \citep{church_2000_empirical}. An LLM could easily predict such repeated entities, even in cases where those are difficult for humans to remember---in a 19th century Russian novel with an extensive cast of characters, for example (see also Figure \ref{fig:issues}).

There is evidence that Transformer LLMs in fact utilize this property of their architecture to make predictions: With sufficient training, they can assign probabilities of near one to sequences of words that they observed earlier \citep{armeni_2022_characterizing}.
This suggests that these models can learn to implement a form of copying mechanism that leverages their decay-free store of the input \citep{olsson_2022_in, bietti_2023_birth, jelassi_2024_repeat}.
Indeed, direct comparisons between humans and LLMs during the processing of texts that contained repeated passages have shown that while mainstream LLMs assigned high probabilities of near one to the repeated portion of the text, humans were not able to predict the repeated words as accurately \citep{vaidya_2023_humans}, and when reading the repeated portion, they did not speed up as drastically as might be expected if they had verbatim memory of the first occurrence of the text \citep{gruteke_2024_effect}.

Mainstream LLMs' perfect store of earlier words in the text input may lead to other discrepancies with human language processing.
One such case could arise when readers form syntactic dependencies between words, for example, when they need to identify the subject for a particular verb from multiple candidate subjects.
In the sentence \textit{The rat the cat the dog chased loved ate the malt}, for example, readers often struggle to determine that the subject of \textit{ate} is \textit{the rat} rather than any of the other noun phrases \citep{gibson_1998_linguistic, mcelree_2003_memory, yngve_1960_model}.
LLM probabilities fail to accurately predict the increase in reading times displayed by participants at the main verb of such deeply embedded sentences \citep{hahn_2022_resource}.
This suggests that Transformers face little interference from multiple candidate subjects, possibly because they can perform many parallel memory retrieval operations at once \citep{timkey_2023_language}.
Interference from multiple candidate subjects in language processing also arises in the phenomenon of agreement attraction \citep{bock_1991_broken, pearlmutter_1999_agreement, wagers_2009_agreement}, where participants, again because they interpret an incorrect noun as the subject of the sentence, consider ungrammatical sentences such as \textit{The keys on the table is rusty} to be acceptable.
Transformer LLMs show a much lower rate of agreement attraction errors than people do, suggesting again that they display considerably less memory interference than humans \citep{arehalli_2024_neural}.

A third mismatch between LLM and human prediction that could be due to superhuman short-term memory concerns syntactic disambiguation.
When a sentence turns out to have a structure that differs from the one it appeared likely to have based on the first few words of the sentence---so called garden-path sentences \citep{bever_1970_cognitive, frazier_1982_making}, such as \textit{The old man the boat}---humans often experience severe processing difficulty \citep{frazier_1982_making, garnsey_1997_contributions, sturt_1999_structural}.
Mainstream LLMs, by contrast, predict only very mild difficulty in those contexts, quite possibly because they have the memory capacity to consider multiple possible structures of the sentence concurrently \citep{vanschijndel_2021_single, arehalli_2022_syntactic, huang_2024_large}.

Further support for the hypothesis that mainstream LLMs' short-term memory is too strong to model human linguistic prediction comes from studies that manipulate or limit LLMs' access to the representations of previous words.
In deeply embedded sentences, downweighting the contribution of high-frequency words that are unlikely to be retained in memory reduces the mismatch between LLMs and humans in the prediction of the main verb \citep{hahn_2022_resource}.
The alignment between LLM probabilities and human reading times improves when the LLM's ability to access earlier words in the sentence is reduced by increasing the attention allocated to recent words \citep[][but cf.~\citealp{thamma_2025_human}]{varda_2024_locally, clark_2025_linear} or providing only a few recent words as input \citep{kuribayashi_2022_context}.
Finally, limiting access to earlier words also yields attention patterns in Transformer LLMs that are better suited to capture processing difficulty during reading \citep{ryu_2025_memory}.
Overall, while questions about the exact form of human-like decay remain, these results converge to the conclusion that mainstream LLMs' access to decay-free representations negatively impacts their ability to model human linguistic prediction.

\section*{Towards LLMs that are better aligned with humans}
How could changes to language model architectures and training procedures address the issues we have identified?
The most obvious discrepancy that needs to be addressed is the mismatch in the quantity and quality of training data. Limiting training data would not only limit LLMs' opportunities to memorize knowledge that humans may not have, but also hinder the development of superhuman short-term memory mechanisms, which emerge in Transformers only after considerable training \citep{armeni_2022_characterizing}.
Such efforts could follow in the footsteps of the BabyLM Challenge \citep{warstadt_2023_proceedings, wilcox_2025_bigger}, a yearly shared task that invites teams to train language models on curated datasets that consist of 100 million words or less, about 40\% of which is transcribed child-directed speech or text material intended for children.
This shared task has shown that language models can learn robust linguistic generalizations from human-scale amounts of data, pointing to the feasibility of this paradigm. But it is as of yet unknown if training on this corpus addresses the superhuman long-term and short-term memory issues described above.

A fundamental question remains about the correct mixture of training data that will give rise to the next-word predictions that are representative of those of the typical human reader.
Identifying the right kind of training data is complicated by the fact that humans learn through other sensory experience in addition to linguistic input \citep{bender_2020_climbing};
for example, most humans might learn the fact that bananas are typically yellow from having seen bananas, rather than from reading text that describes the color of bananas. This raises the possibility that fully human-like word predictions would require training on multimodal data, or additional textual data that would compensate for the lack of multimodal training.

Even with the ideal training data, the misalignment will persist if models draw different generalizations from the data than humans.
The typical training objective of LLMs only rewards predicting the correct word---the one that in fact occurred in the text---and consequently mainstream LLMs are incentivized to make lexically sharper predictions that assign a high probability to the correct continuation whenever possible.
For example, given \textit{Elvis Presley was born in the city of} \underline{\hspace{1cm}}~, they are likely to assign a high probability to \textit{Tupelo} and lower probabilities to names of other cities.
Given mainstream LLMs' ability to store training examples, these lexically sharp predictions are unlikely to change once they are learned.
In contrast, humans make broader predictions based on meaning. These predictions facilitate the processing of words that share semantic properties with the target word, even if the target word itself is unpredictable \citep{federmeier_1999_rose, roland_2012_semantic, luke_2016_limits}.
Therefore, a reader who does not know the birthplace of Elvis is likely to find the name of a plausible yet incorrect city to be somewhat predictable.
While we expect that in such cases LLMs also assign some probability mass to words other than the most likely word, we hypothesize that they do so to a lesser extent than people.

To incentivize LLMs to make next-word predictions that are as diffuse as those of humans, alternative training objectives that upweight semantically similar words together could be helpful.
Such a training objective could minimize a continuous loss based on the distance between vector representations \citep{kumar_2019_von}, which would upweight the prediction of not only \textit{Tupelo} but also other cities, to the extent that their representations are similar.
This can be coupled with a reduction in the size of vector representations for each word, which would further prevent the model from encoding specific information about each city.
Another way to make LLMs' predictions more diffuse could involve temperature-scaling prior to probability calculation to increase the uncertainty of LLMs' predictions \citep{gao_2017_calibration}; this intervention has been shown to improve the fit of LLM probabilities to human data \citep{liu_2024_temperature}.

The misalignment in predictions that is due to mainstream LLMs' long-term memory can also be addressed post-hoc by intervening on their representations---first locating a piece of information of interest, and then adjusting it to steer the model's predictions.
Model editing techniques \citep{meng_2022_locating, wang_2023_knowledge, cohen_2024_evaluating}, in particular, typically train the model on a curated set of examples so as to lead it to make predictions that are more factually correct than it would otherwise.
For instance, if a model mispredicts the birthplace of Elvis Presley as \textit{Orlando}, one could further train it on sentences like \textit{Elvis Presley was born in Tupelo} to get it to predict the correct birthplace of Elvis. 
For purposes of cognitive modeling, these techniques can be applied to cause it to unlearn knowledge that most human readers wouldn't have.

What about LLMs' superhuman short-term memory?
Existing approaches for reducing the accessibility of the representations of earlier words in Transformer LLMs make simplifying assumptions about human memory---in particular, that the input decays solely as a function of time, such that earlier words in the text always decay more than later words \citep{kuribayashi_2022_context, varda_2024_locally, clark_2025_linear}, or as a function of frequency, such that high-frequency words always decay more than low-frequency words \citep{hahn_2022_resource}.
As a starting point, experiments should evaluate the impact of more flexible methods for limiting Transformers' memory retrieval accuracy.
Those could include decay at the level of syntactic constituents rather than individual tokens \citep{lewis_2005_activation}, or memory retrieval bottlenecks that could give rise to the interference humans experience between similar units in short-term memory \citep{timkey_2023_language}.

\begin{figure}[t!]
\vspace{\baselineskip}
\begin{tcolorbox}[title=Box 1.~Targeted human experiments for benchmarking modeling progress, fontupper=\linespread{1.25}\selectfont]
    We have outlined the hypothesis that the discrepancy between the linguistic predictions of mainstream LLMs and humans is rooted in the models' superior long-term and short-term memory.
    Existing evidence for this hypothesis is observational and preliminary, as many of the reading datasets that have been studied focus on naturalistic text (e.g.~newspaper articles), and as such do not manipulate short-term and long-term memory in a controlled way.
    To better ground this hypothesis, and to benchmark efforts to close the gap between LLMs and humans, new, targeted human experiments are necessary.

    \hspace{15pt} For the first issue surrounding long-term memory, a more direct link between the readers' knowledge of text and their reading behavior needs to be established.
    For the example of factual knowledge, this could be achieved by collecting reading times of sentences from datasets that contain information about real-world entities \citep[e.g.][]{levy_2017_zero}, coupled with a questionnaire that gauges familiarity with the relevant piece of knowledge.
    A similar experimental paradigm, which  measures humans' familiarity with some text and links it to their reading behavior, can likewise apply to multiword expressions and famous documents like the Declaration of Independence.

    \hspace{15pt} For the second issue of short-term memory, the influence of temporal decay in reading---which needs to be evaluated at the discourse level---has been far less studied than the influence of complex subject-verb relations within the same sentence.
    As such, we recommend collecting measurements during the reading of texts that implicitly require readers to recall information, such as a short list of items or the name of a particular character in the story. These experiments should vary factors that are known to affect human memory, such as the size of the list, the amount of intervening text between mentions, and the position of the entity in the first presentation of the list \citep{oberauer_2018_benchmarks}.
\end{tcolorbox}
\end{figure}

Alongside these approaches, which aim to restrict the scope of Transformers' attention mechanisms, we recommend exploring approaches that abandon Transformers altogether, and place a renewed focus on recurrent neural network architectures \citep[e.g.][]{elman_1991_distributed, hochreiter_1997_long, dao_2024_transformers}.
In comparison to Transformers, which have direct access to the representations of prior words, recurrent models are forced to compress the input into a representation of a fixed size, and therefore are more likely to implement a human-like form of decay when making predictions. 

\begin{figure}[t!]
\vspace{\baselineskip}
\begin{tcolorbox}[title=Outstanding questions, fontupper=\linespread{1.25}\selectfont]

\begin{itemize}[leftmargin=11pt]
    \item What kinds of training data and training objectives are most likely to give rise to next-word predictions that average human readers make?
    \item How does extralinguistic experience---to which many LLMs typically do not have exposure---shape humans' linguistic predictions during reading?
    \item Compared to LLMs, how much linguistic and non-linguistic knowledge do average human readers lack? Conversely, is there knowledge that humans have but LLMs do not that also introduces a mismatch in predictions?
    \item What are the most effective machine learning techniques for erasing the knowledge encoded by LLMs, without drastically distorting their representations and predictions?
    \item What kinds of linguistic information are retained in human short-term memory, and how can neural network architectures be modified to match this mechanism?
\end{itemize}
\end{tcolorbox}
\end{figure}

These efforts to narrow the gap between LLMs and humans should be accompanied by targeted human experiments that measure the role of long-term and short-term memory in human reading (Box 1).
Such experiments will help benchmark research progress on this front and complement existing datasets in providing a more detailed quantitative characterization of the misalignment between the memory capabilities of humans and LLMs as it affects linguistic prediction.

\section*{Concluding remarks}
Prediction is a core principle of human cognition and language processing: We use our linguistic experience to make predictions about how a sentence will unfold.
While language models could help elucidate this process, in the last few years mainstream LLMs have become much more accurate than humans at next-word prediction, limiting their applicability for cognitive modeling.
This discrepancy between humans and LLMs partly parallels findings in the vision domain, where deep neural networks use prediction strategies that exceed humans' capabilities \citep{linsley_2023_performance, linsley_2025_better}.
We highlighted two areas where LLMs are superhuman in this sense: their superior long-term memory of training examples, and their superior short-term memory of previous words in the text.
We call for research effort to address these issues (see also \textit{Outstanding questions}): We should develop language models with human-like memory through alternative training procedures and neural network architectures, and conduct new human experiments to benchmark research progress.

\section*{Acknowledgments}
This project is supported by the National Science Foundation (NSF) under grants BCS-2020945 and IIS-2504953 and the National Institute of Biomedical Imaging and Bioengineering under grant R01EB038873.
The content is solely the responsibility of the authors and does not necessarily represent the official views of the National Science Foundation or the National Institutes of Health.

\section*{Declaration of Interests}
Tal Linzen is a part-time employee of Google.

% ====================

\bibliographystyle{myapalike}
\bibliography{mybib}

% ====================

\end{document}